\documentclass[letterpaper, 10 pt, conference]{ieeeconf}  

\IEEEoverridecommandlockouts                              

\usepackage{graphics} 
\usepackage{epsfig} 
\usepackage{mathptmx} 
\usepackage{times} 
\usepackage{amsmath} 
\usepackage{amssymb}  

\usepackage{gensymb}
\usepackage{booktabs}
\usepackage{multirow}
\usepackage{caption} 
\usepackage{subfigure}
\usepackage{outlines}
\usepackage{lipsum}
\usepackage{xspace}
\usepackage{comment}
\usepackage{pbox}    
\usepackage{pifont}
\newcommand{\cmark}{\ding{51}}%
\newcommand{\xmark}{\ding{55}}%

\usepackage{xcolor}

\newcommand{\etal}{\textit{et al.}}
\newcommand{\ie}{\textit{i.e.}}
\newcommand{\eg}{\textit{e.g.}}

\usepackage{xcolor}

\usepackage{cuted}

\begin{document}

\title{\LARGE \bf From 2D to 3D: Re-thinking Benchmarking of \\ Monocular Depth Prediction}

\author{Evin P{\i}nar \"{O}rnek$^{1,*}$, Shristi Mudgal$^{1,*}$, Johanna Wald$^{1,2}$, Yida Wang$^{1}$, Nassir Navab$^{1}$ and Federico Tombari$^{1, 2}$
\thanks{$^1$:Technical University of Munich, Germany; {\tt\small \{evin.oernek, shristi.mudgal, yida.wang, nassir.navab, federico.tombari\}@tum.de}}
\thanks{ $^2$:Google Inc. {\tt\small johannawald@google.com}}
\thanks{$*$:the first two authors contributed equally.} }

\maketitle
\noindent
\begin{strip}
\vspace{-2.5cm}
    \centering
    \captionsetup{type=figure}
	\includegraphics[width=\linewidth]{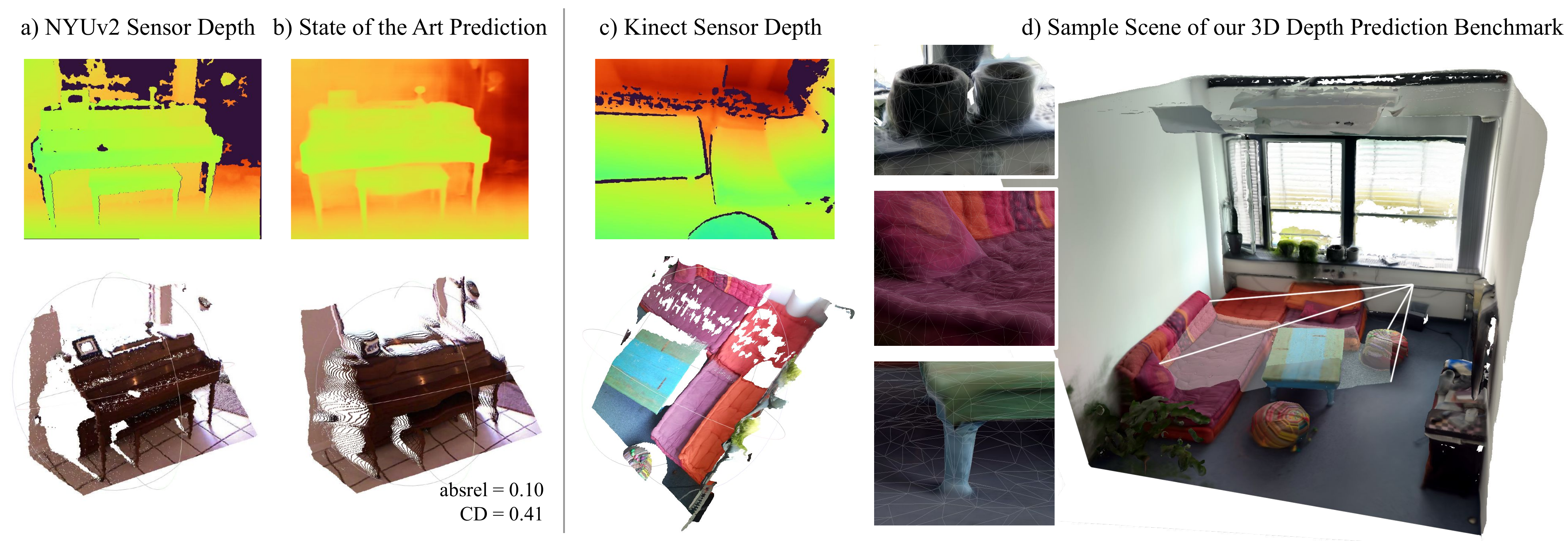}
   \caption{State of the Art depth prediction with low absolute relative error (absrel) often does not exhibit correct 3D structures when back-projected in 3D space (b). We suggest using 3D metrics, \eg, chamfer distance (CD), to measure the structural quality of the prediction and propose a new depth prediction benchmark (d) with entire 3D geometry and high-quality depth compared to raw sensor measurements (a, c). Notice invalid depth regions of the sensor depth maps shown in black.}
   \label{fig:teaser}
\end{strip}

\thispagestyle{empty}
\pagestyle{empty}


\begin{abstract}

There have been numerous recently proposed methods for monocular depth prediction (MDP) coupled with the equally rapid evolution of benchmarking tools. However, we argue that MDP is currently witnessing benchmark over-fitting and relying on metrics that are only partially helpful to gauge the usefulness of the predictions for 3D applications.
This limits the design and development of novel methods that are truly aware of - and improving towards estimating - the 3D structure of the scene rather than optimizing 2D-based distances. In this work, we aim to bring structural awareness to MDP, an inherently 3D task, by exhibiting the limits of evaluation metrics towards assessing the quality of the 3D geometry. 
We propose a set of metrics well suited to evaluate the 3D geometry of MDP approaches and a novel indoor benchmark, RIO-D3D, crucial for the proposed evaluation methodology. Our benchmark is based on a real-world dataset featuring high-quality rendered depth maps obtained from RGB-D reconstructions. We further demonstrate this to help benchmark the closely-tied task of 3D scene completion.

\end{abstract}

\section{INTRODUCTION}

Recent computer vision literature has witnessed an abundance of research on monocular depth prediction (MDP), \ie, the task of predicting a dense depth map from a single RGB image. Compared to traditional approaches~\cite{hoiem2005,saxena2009} which were proposed more than a decade ago, the advent of deep learning in the last few years offers more effective ways~\cite{eigen2014,laina2016,ramamonjisoa2019sharpnet,lee2019bts} to regress depth maps if a vast set of data is available for training. However, the literature is trending towards over-fitting the development of novel ideas to the maximization of a small subset of specific metrics on these datasets. This poses a limit to the development of more precise and more useful MDP approaches for two reasons: i) the main metrics tend to disregard the accuracy for the 3D structure of the predicted geometry, so the depth maps scoring high values for these metrics do not reflect correct 3D structures when they are back-projected in 3D space (see Fig.~\ref{fig:teaser}(a, b)); current metrics compute the accuracy globally and up to the 2D depth plane rather than in the 3D domain ii) some datasets are supplemented with noisy 3D supervisions which are represented by the raw depth data obtained from RGB-D sensors, such that training and evaluating will be biased by noise in the ground truth. This is portrayed by the examples shown in Fig.
~\ref{fig:teaser}(a, c). We argue that improving 3D accuracy of predictions is of utmost importance for MDP -- since the vast majority of applications using depth prediction relies on 3D geometry of the scene and the objects therein -- in order to, e.g., navigate unseen environments through obstacle avoidance, grasp objects within clutter or augment shapes with 3D information. However, the accuracy of predicted 3D geometry, is not sufficiently represented in the commonly used 2D-based metrics nor in the network designs and loss functions being proposed. This results in continual research in a direction that does not allow the true growth of MDP in terms of the quality of predictions. 

Based on this preamble, the goal of our work is to move beyond this local minimum by exploring novel avenues to develop, evaluate and benchmark MDP in a more \textit{3D-driven} way, such that its applications could be well served. Furthermore, we noticed that the current MDP datasets are not supplemented with the necessary size, quality or semantics, and a 3D-based evaluation requires depth maps of high quality on 3D space. Hence, we propose a large-scale indoor dataset RIO-D3D (see Fig.~\ref{fig:teaser}(d))) where aside from semantic information, a high quality, complete and denoised 3D ground truth is available, resulting in a cleaner and finer structured, dense point cloud. Our dataset also enables evaluating the task of semantic scene completion, which is directly related to MDP, as both aim towards a full 3D reconstruction of a scene. The proposed benchmark dataset and evaluation codes will be publicly available.

In summary, our contributions are the following:
\begin{enumerate}
    \item We propose a comprehensive evaluation protocol based on 3D metrics, aiming at assessing the prediction quality in terms of 3D geometry.
    \item A novel large-scale, real, indoor benchmark (see Fig.~\ref{fig:teaser}(d)) dataset RIO-D3D is provided for MDP with highly accurate, complete 3D ground truth, together with normal, occlusion and semantic annotations; allowing training and evaluation of more suited, 3D aware approaches.
    \item We provide a comparison among state of the art MDP and 3D scene completion methods along with the proposed evaluation criteria on RIO-D3D dataset. 
    \item We bring to light the need to shift the focus of MDP research towards the structural information in the scene and better quantify the prediction quality. This is affirmed by experimental results, highlighting the flaws of 2D metrics and 2D-based approaches.
\end{enumerate}

\section{RELATED WORK}

\noindent\textbf{Monocular depth prediction} 
Initially, traditional works on MDP~\cite{hoiem2005,liu2011sift,oliva2001hog,saxena2005,karsch2012} build the inference models based on hand-crafted features tied to the depth priors in the scenes. Deep learning provides more potential when large scale training data is available, the very first approach~\cite{eigen2014} proposed is based on AlexNet~\cite{alexnet2012}/VGG~\cite{Simonyan15}. \cite{eigen2015}~improved the performance by stacking more layers and predicting the normals and semantic labels. Laina~\etal~\cite{laina2016} introduced a ResNet-based fully convolutional network with novel up-sampling blocks while other works~\cite{zhuo2015,li2015,wang2015} combined multi-scale convolutional neural network features with a Conditional Random Field refinement. Several methods attempted to model MDP as a relative ordering task. Fu~\etal~\cite{fu2018dorn} applied ordinal regression, Dhamo~\etal~\cite{dhamo2019} used a layer-wise scene modeling, Lee \etal~\cite{lee2019relative} proposed a model for ordinal consistency. 
The heavy-tailed distribution in depth maps, where large amounts of pixels are predominantly small depth values, leads to data imbalance problem. Reversed Huber loss~\cite{laina2016}, binary decision trees \cite{roy2016}, attention weighted loss~\cite{jiao2018deeper} were proposed for this. Recently, Bhat \etal~\cite{Bhat2021AdaBinsDE} proposed a novel formulation for predicting distance values by means of classification. Tian \etal~\cite{Tian2020} proposed attention blocks within the decoder, while other methods adopt a Transformer based architecture~\cite{Ranftl2021,yang2021transformers}.

Further, multi-task learning is being actively explored, focusing on jointly learning depth prediction with tasks such as surface normals or semantic segmentation.
Some of the works \cite{li2017twostream,jafari17,mousavian2016} aim at learning a globally consistent depth map but do not focus on fine local details, often resulting in blurry boundaries and deformed planar surfaces.
Several works \cite{hu2018boundary,shih2020,niklaus2019,ramamon2020,cliffnet2020} focus on accurate object boundaries. Ramamonjisoa~\etal~\cite{ramamonjisoa2019sharpnet} aim to improve the edges by predicting normals and edges along with depth.   
In general, multi-task methods need complementary data 
in addition to the depth ground truth which is often missing. Our proposed benchmark provides semantics, normal maps, as well as occlusion boundaries for the entire dataset.

\noindent\textbf{3D-aware depth prediction}
Most of the aforementioned methods use 2D networks and predict depth using a structured 2D pixel grid. These works are trending towards over-fitting to a subset of commonly used 2D metrics. Only recently have approaches been proposed that bring more 3D awareness to MDP. Jointly estimating 3D surface normals~\cite{zhang2019pap,lee2019bts} or enforcing planar constraints ~\cite{liu2019planercnn} improves depth predictions in indoor setups. Liu~\etal reconstruct a scene from an arbitrary number of piece-wise planar surfaces~\cite{liu2019planercnn}. The first work to highlight the role of depth maps in understanding 3D scene geometry is Yin \etal \cite{yin2019virtualnormal}. They introduce a 3D training loss which backprojects the depth to 3D and enforces consistency on 3D surface patches. While these methods present convincing performance, they are not only limited by the quality of the training data but also by the metrics used when evaluating.

\begin{table}[t!]
\caption{Comparison of indoor depth prediction datasets.}
\begin{center}
\resizebox{\columnwidth}{!}{%
\begin{tabular}{lccccccc}
\textbf{Dataset} & \textbf{Sem.} & \textbf{Full 3D} & \textbf{Setting} & \textbf{Acquisition} & \textbf{Size} \\
\hline
NYU v2 \cite{silberman2012} & \cmark & \xmark & indoor & Kinect & 1449 \\
SUN RGB-D \cite{song2015sunrgbd} & \cmark & \cmark & indoor & various& 10K \\
DIW \cite{diw2016}  & \xmark & \xmark & both & web (relative) & 500K \\
Matterport \cite{Matterport3D} & \cmark &  \cmark & indoor & Matterport & 200K \\
2D-3D-S \cite{armeni2017joint} & \cmark & \cmark & indoor & Matterport & 70K \\
ScanNet \cite{dai2017scannet} & \cmark & \cmark & indoor &  structured & 24K \\
ETH 3D \cite{eth3d2019}  & \xmark & \xmark & both & laser & 545 \\
DIODE \cite{diode_dataset2019} & \xmark & \xmark  & both & Faro & 27K \\
\hline
RIO-D3D & \cmark & \cmark & indoor & 3D scan  & 34K

\end{tabular}
}
\end{center}
\label{tbl:datasets}
\end{table}

\noindent\textbf{Datasets and metrics} Numerous datasets have been introduced for data-driven deep learning models, see Tbl.~\ref{tbl:datasets}. 
The earlier benchmarks are afflicted by either low resolution, sparse depth, small size and/or sensor limitations of the time. 
NYUv2~\cite{silberman2012} is by far the most popular indoor benchmark featuring real-world environments but includes only a small number of semantic images, inline with other datasets that were later introduced with semantic annotations~\cite{song2015sunrgbd,armeni2017joint,Matterport3D}. 
ScanNet~\cite{dai2017scannet} while providing 3D reconstructions, does not provide 3D reconstructed depth maps. It is a sensor based depth benchmark with 24K images which, due to the sensor, similarly fails on low albedo and reflective surfaces. 
As opposed to these works, ETH~\cite{eth3d2019} or DIODE~\cite{diode_dataset2019} provide finer ground truth but lack semantics and availability of the full 3D scene including occluded regions (see Fig.~\ref{fig:teaser}(d)) in addition to DIODE being relatively small. 

The evaluation on these datasets is commonly carried out by using a set of metrics: absolute relative distance (Eq. \ref{eq:rel}), squared relative distance (Eq. \ref{eq:sqrel}), root mean square error (Eq. \ref{eq:rmse}), logged root mean square error (Eq. \ref{eq:logrmse}) and accuracy under threshold (Eq. \ref{eq:acc}).
\begin{equation} \label{eq:rel}
 \text{absrel} = \frac{1}{|T|}\sum_{y \in T }\frac{|y - y^{*}|}{y^{*}},
\end{equation}
\begin{equation} \label{eq:sqrel}
 \text{sqrel} = \frac{1}{|T|}\sum_{y \in T }\frac{{\|y - y^{*}\|}^{2}}{y^{*}},
\end{equation}
\begin{equation} \label{eq:rmse}
    \text{rmse (linear)} = \sqrt{
\frac{1}{|T|} 
\sum_{y \in T }
{\|y - y^{*}\|}^{2}
},
\end{equation}
\begin{equation} \label{eq:logrmse}
\text{rmse (log)} = \sqrt{ 
\frac{1}{|T|} 
\sum_{y \in T }
{\|\log y - \log y^{*}\|}^{2}
},
\end{equation}
\begin{equation} \label{eq:acc}
 \text{accuracy at $\delta$,} = \% \text{ of } y_{i} \text{ s.t. } max(\frac{y_{i}}{y_{i}^{*}}, \frac{y_{i}^{*}}{y_{i}}) = \delta < t.
\end{equation}

\noindent Given a ground truth depth map and the corresponding prediction, metrics are calculated iterating over all pixels in dataset $T$, comparing the ground truth value $y^*$ with its predicted counterpart $y$, and averaging over all data samples $|T|$.

In addition to standard 2D metrics, Koch~\etal~\cite{koch2018} recently suggested an occlusion boundary metric ${acc}_{DBE}$ to measure the occlusion boundaries on depth maps. Furthermore, they propose plane accuracy metric to measure the connectivity and geometry of planar surfaces. Inspired from them, Ramamonjisoa \etal \cite{ramamonjisoa2019sharpnet} also emphasized the occlusion boundary metric. However, none of these works evaluate depth maps in terms of 3D structures. To the best of our knowledge, our work is the first to explore the usage of different 3D metrics when evaluating MDP.

\section{EVALUATION METHODOLOGY}

Figure \ref{fig:teaser}(b) shows the predicted point cloud obtained using an available pre-trained model. We can observe consistent deformations and noise, especially around the occlusion boundaries. This can be attributed to the 2D-oriented network architectures, datasets, 2D depth value based loss functions, and deficient evaluation criteria. This work addresses two of these aspects: evaluation criteria and datasets.

To that end, we redefine MDP as a single-view 3D reconstruction task. Inspired by the related 2D-to-3D tasks, we propose the usage of 3D metrics and losses to complete the missing 3D aspects within MDP. While most of these metrics have already been proposed earlier~\cite{fan2017,tatarchenko2019}, we are the first to apply them when evaluating depth prediction.

\paragraph{Earth Mover's Distance (EMD)}  

Consider two point clouds $G, R \subseteq \mathbb{R}^{3}$ of equal size $|G|=|R|$, where $G$ is the ground truth, $R$ is the reconstruction. EMD between $G$ and $R$ is defined as follows:
\begin{equation} 
    d_{EMD}(G, R)=\min_{\phi:G\rightarrow R} \sum_{x\in G} \|x-\phi(x)\|_2,
\end{equation}
\noindent where $\phi:G\rightarrow R$ is a bijection. Practically, finding this bijection mapping is too expensive and we follow an approximate implementation as in \cite{fan2017}. 

\paragraph{3D Chamfer Distance (CD)} $d_{CD}(G, R)$ between $G, R\subseteq \mathbb{R}^3$ is defined as:

\begin{equation} \displaystyle
    d_{CD}(G, R) = \sum_{x\in G}\min_{y\in R} \|x-y\|^2_2+\sum_{y\in R}\min_{x\in G} \|x-y\|^2_2,
\end{equation}
 \noindent where each point from $G$ is matched with the nearest neighboring point from $R$ and vice versa. CD is known to be sensitive to outliers and potentially finds wrong correspondences when points are incorrectly matched. 

\paragraph{3D Completeness} 
This score closes the gap between EMD and CD. EMD is typically calculated from ground truth to prediction, whereas CD covers both directions simultaneously. 
Completeness measures the distances from $G$ to $R$, showing the error on the retrieved points. It can be calculated as:
\begin{equation} \displaystyle
    d_{EMD}(R, G)=\min_{\phi:R\rightarrow G} \sum_{x\in R} \|x-\phi(x)\|_2,
\end{equation}
Completeness is necessary to understand whether the reconstructed points lie within an acceptable the range of ground truth points.

\begin{figure*}[t!]
    \centering
    \includegraphics[width=0.9\linewidth]{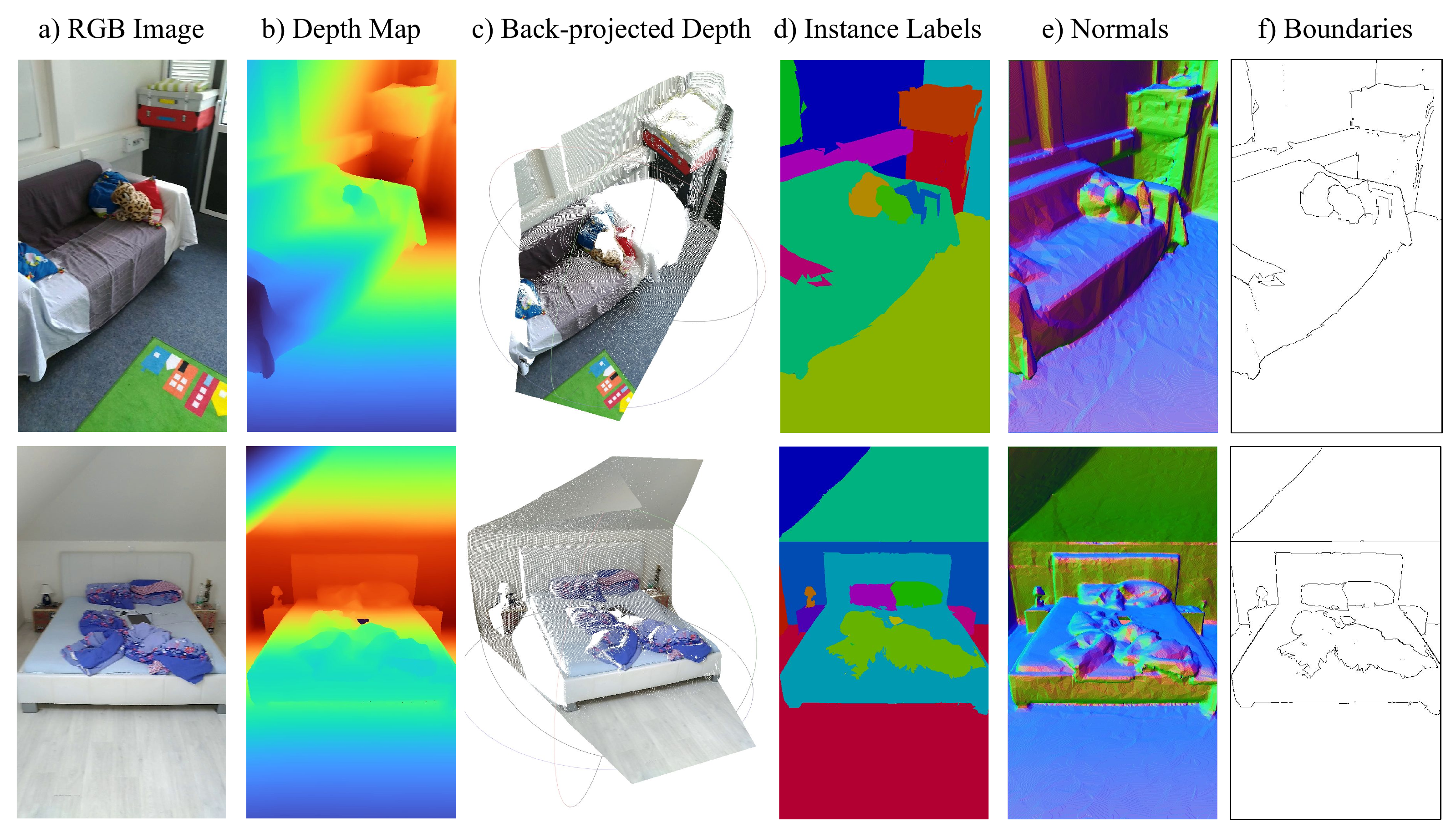}
   \caption{Two samples from RIO-D3D a) RGB Images, corresponding b) depth map (red is far, blue is close), c) back-projected depth, d) instance segmentation, e) normals, and f) instance and occlusion boundaries.}
\label{fig:dataset_examples}
\end{figure*}

\paragraph{3D Point Cloud F-score} F-score, also known as Dice coefficient, is the harmonic mean of precision and recall. Precision is the number of predicted points lying within a distance to a ground truth surface, whereas recall is the number of ground truth points lying within the prediction. The metric was introduced by Knapitsch~\cite{knapitsch2017} to evaluate 3D surface quality, and further emphasized by Tatarchenko \etal \cite{tatarchenko2019}.

\noindent The recall with a distance threshold $t$ is defined as: 

\begin{equation} \displaystyle
    R(t) = \frac{1}{|G|} \sum_{g \in G}[e_{g\rightarrow R} < t].
\end{equation}
\noindent where the distance $e$ from a ground truth point $g$ to a reconstructed point $r$ is $e_{g\rightarrow R} = \min_{r \in R} \| g - r\|.$

\noindent Likewise, the precision is:
\begin{equation} \displaystyle
    P(t) = \frac{1}{|R|} \sum_{r \in R}[e_{r\rightarrow G} < t].
\end{equation}

\noindent where the distance $e$ from a reconstructed point $r \in R$ to ground truth $G$ is defined as $e_{r\rightarrow G} = \min_{g \in G} \| r - g\|.$

\begin{equation} \displaystyle
    F(t) = 2 \cdot \frac{P(d) \cdot R(d)}{ P(d) + R(d)}.
\end{equation}

\paragraph{3D Point Cloud Intersection Over Union} IoU is commonly used to measure the volumetric quality of a 3D reconstruction. It is also known as Jaccard Index calculated between two input sets. 
 \noindent 3D IoU in literature is computed on voxelized point clouds, which requires a sampling and discretization step, losing the ability to reflect surface quality. We instead propose an IoU calculation on non-voxelized point clouds. Inspired by the generalized formulation of the F-score, Jaccard and the Tversky index \cite{tversky77}, and their relationship formulated as $\frac{J}{F} = \frac{1}{2} + \frac{J}{2}$, we re-express the point based IoU in terms of precision and recall:

\begin{equation} 
    IoU(d) = \frac{P(d) \cdot R(d)}{ P(d) + R(d) - P(d) \cdot R(d)}.
\end{equation}

We compare these 3D-based metrics on the proposed dataset to evaluate MDP in the experiments section. 

\section{EVALUATION DATASET: RIO-D3D}
\label{sec:dataset}

We propose RIO-D3D, a new large-scale, highly variable, an indoor dataset with calibrated RGB-D sequences, high-quality 3D ground truth, instance-level semantic annotations, associated normals, and geometric as well as occlusion boundaries. We build upon the 3RScan dataset~\cite{wald2019rio} which features 1482 RGB-D scans of 478 unique environments, camera poses, and intrinsic parameters. They processed each sequence offline to get bundle-adjusted camera poses with loop-closure and texture-mapped 3D reconstructions. The original dataset has 363K real images, yet, the provided 3D reconstruction allows the generation of unlimited semi-synthetic data from various viewpoints to create additional training data. 

The dense 3D reconstruction reduces missing surfaces and mitigates noise (particularly on edges and in low lighting environments) typically present in sensor-acquired depth maps. We render depth maps associated with the RGB frames using the camera poses and the intrinsic data allowing access to a more complete depth map with higher coverage and rich 3D structural information. This results in a ground truth of superior quality than common sensors, making the benchmark suitable for training and evaluating different kinds of approaches. 

A good depth prediction dataset should provide not only RGB-depth pairs in a large-scale setting but also semantic and additional geometric information to help propel multi-task learning.
When compared to the existing datasets, RIO-D3D provides not just high-quality ground truth but also normal maps, occlusion boundaries, semantic and instance labels for the entirety of our dataset. The normal maps and occlusion boundaries are generated by following a point cloud-based approach from Tateno \etal \cite{tateno2017cvpr}. 


\begin{figure}[t]
    \centering
        \subfigure[Depth dist.] {
        \includegraphics[width=0.43\linewidth]{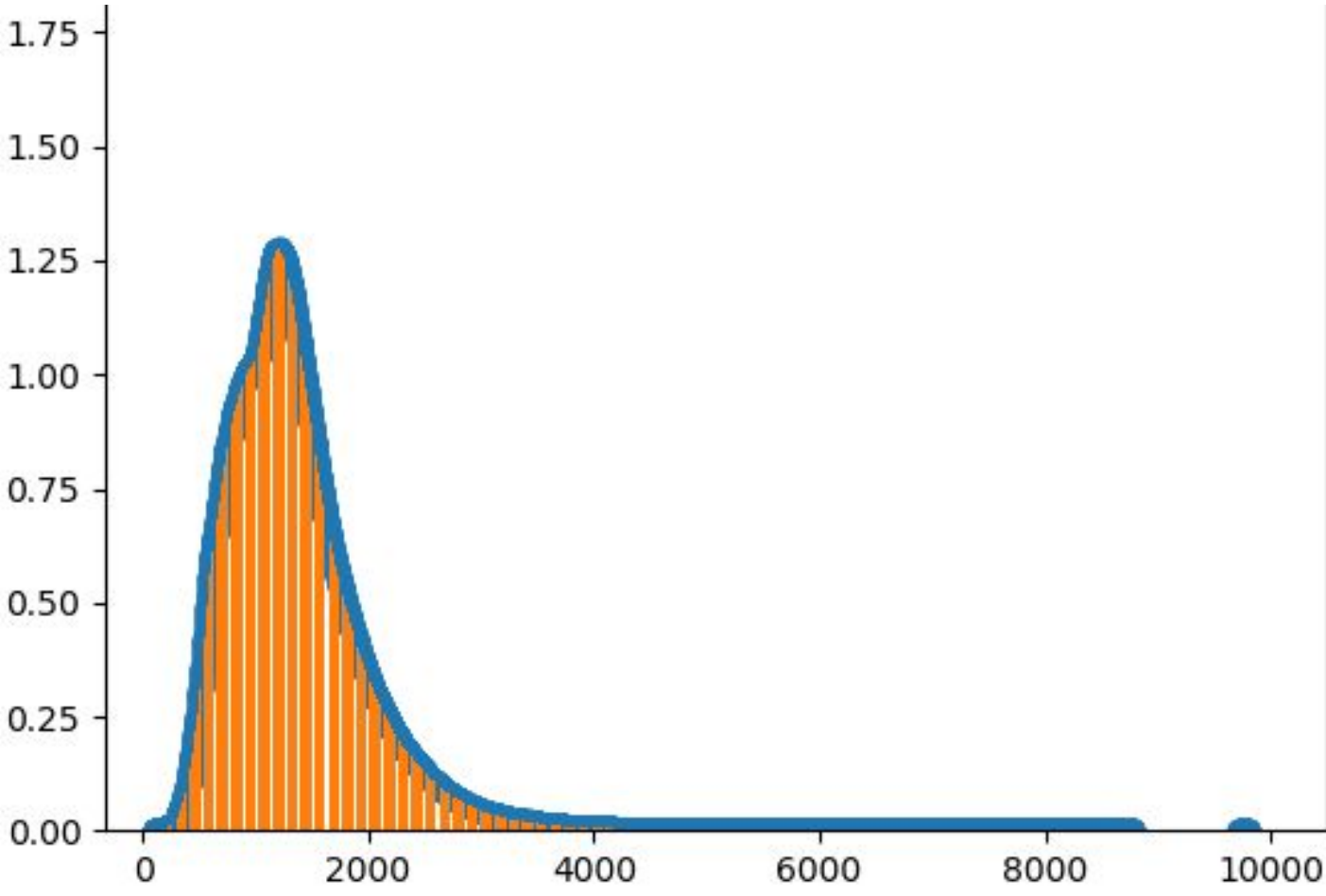}}
        \subfigure[Coverage] {
        \includegraphics[width=0.43\linewidth]{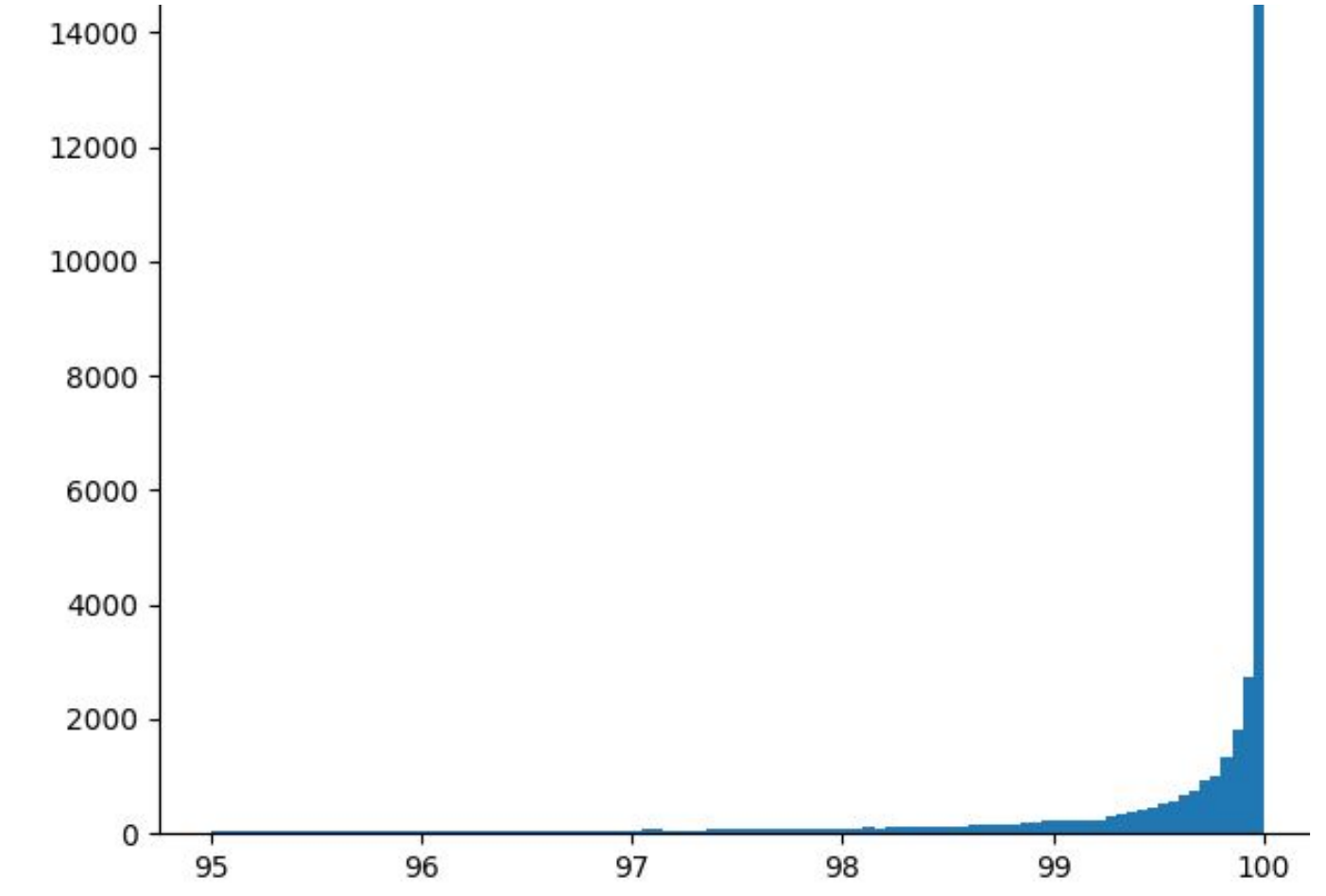}}
    \caption{Depth distribution (in mm over number of pixels) and dataset coverage are shown. }
    \label{fig:depthdist}
\end{figure}

\begin{table*}[t!]
  \centering 
  \begin{tabular}{ l | c c c | c c c c c }
  \toprule
    \multirow{2}{*}{\bfseries Method} & \multicolumn{3}{c}{\bfseries 2D Metrics} & \multicolumn{5}{c}{\bfseries 3D Metrics} \\  & absrel$\downarrow$ & rmse$\downarrow$ & $\delta_{1}\uparrow$ & CD$\downarrow$ & EMD$\downarrow$ & Comp.$\downarrow$ & IoU$\uparrow$ & F-score$\uparrow$ \\
    \midrule
      Oracle NN   & \textbf{0.097} & \textbf{0.225} & \textbf{0.891} & \textbf{0.257} & \textbf{0.130} & 0.128 &  0.328 & 0.335 \\
      Median Plane   & 0.211 & 0.577 & 0.668 & 0.677 & 0.636 & \textbf{0.042} &  \textbf{0.347} & 0.369 \\
    \midrule
      Eigen~\cite{eigen2014} &  0.217 & 0.712 & 0.637 & 0.584 & 0.529 & 0.055  & 0.254 & 0.405 \\
      FCRN~\cite{laina2016}  &  0.217 & 0.703 & 0.647 & 0.491 & 0.430 & 0.061 &  0.273 & 0.428 \\
      BTS~\cite{lee2019bts}   &  0.190  & 0.657 & 0.694 & 0.454 & 0.400 & 0.055   & 0.336 & \textbf{0.500} \\
      VNL~\cite{yin2019virtualnormal}   & 0.258 & 0.638 & 0.534 & 0.764 & 0.686 & 0.078  & 0.219 & 0.313 \\
    \bottomrule 
  \end{tabular}
  \caption[State-of-the-art comparison on metrics]
  {Quantitative comparison of MDP methods on standard 2D-based metrics and the proposed 3D-based metrics. The first two lines show the metric results for Oracle Nearest Neighbour and Median Plane experiments.} \label{tab:metrics}
\end{table*}

\noindent \textbf{Dataset details}
We focus on the real images in the experiments to provide a fair comparison to the state of the art. After preprocessing, we selected a benchmark of ~34K images (8:1:1 train, val and test). This is to provide a content-rich subset with minimal redundancy, considering that larger datasets are often not used in their entirety due to the limitation of train time, for example, the KITTI-Eigen split (24K images) from the full KITTI dataset. 
Depth distribution and coverage statistics can be seen in Fig. \ref{fig:depthdist}.

\section{EXPERIMENTS}

This section will first lay the groundwork to establish the need for a new evaluation protocol through two toy experiments. Then we give the qualitative and quantitative results for our MDP and scene completion benchmarks with the RIO-D3D and proposed evaluation methodology. 

\subsection{What do deep learning based MDP networks learn?}

We carry out two experiments to examine the behavior of MDP under a hypothetical setup and gain insight into how they compare to the performance of existing methods: 

\begin{figure}[t]
    \centering
        \includegraphics[width=0.9\linewidth]{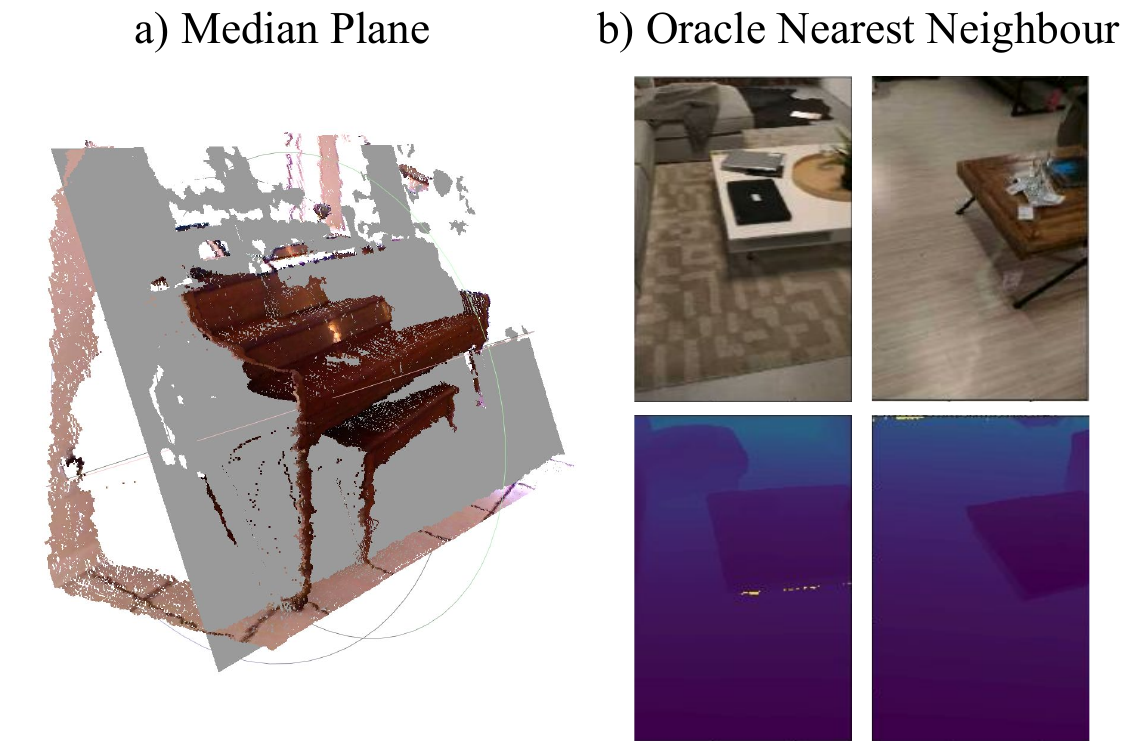}
    \caption{Sample (a) Median Plane (b) Oracle Nearest Neighbour test image (top-left), and corresponding retrieved closest train image (top-right) with their depth maps (below).}
    \label{fig:whatdomdp}
\end{figure}

\noindent\textbf{Median plane} We argue that the existing MDP methods learn an average planar depth due to using 2D-based distances and a global averaging. To test this, we handcraft a set of depth maps, where the distance values are set to the ground truth depth map's median distance value, visualized in Fig. \ref{fig:whatdomdp}(a). We use these planar depth maps as predictions and evaluate them with both 2D and 3D-based metrics. These handcrafted depth maps gave an absrel of $0.211$, which is on par with the model predictions, as shown in Tbl.~\ref{tab:metrics}. This confirmed that if a network were to simply learn to associate median depth values and minimize the loss instead of trying to learn structures, they would still get reasonable 2D metrics results. However, they would struggle on the 3D-based metrics (CD of $0.677$ and F-score of $0.369$).

\noindent\textbf{Oracle nearest neighbour} This experiment measures the performance of a retrieval-driven MDP. How would a method perform that, instead of regressing the distance values, retrieves the structurally nearest image in the training set? To confirm this, for each validation image, we retrieved the nearest depth map from the train split according to the absrel value (see Fig. \ref{fig:whatdomdp}(b)). We then calculated the overall metrics by assuming the nearest depth maps as predictions for the images on the validation set. Results (Tbl. \ref{tab:metrics}) show that OracleNN is indeed superior to the predictions, so retrieving a depth map from the training set gives better quantitative performance than existing regression methods.

These analysis experiments show that MDP methods are not performing as well as they should. We do not see a true improvement in terms of the structure and geometry of the predictions. 
It raises the fundamental question of what networks can truly learn if they are correctly evaluated, and we should re-think how MDP can be better benchmarked. 

\begin{figure*}[t]
\centering
    \includegraphics[width=0.9\linewidth]{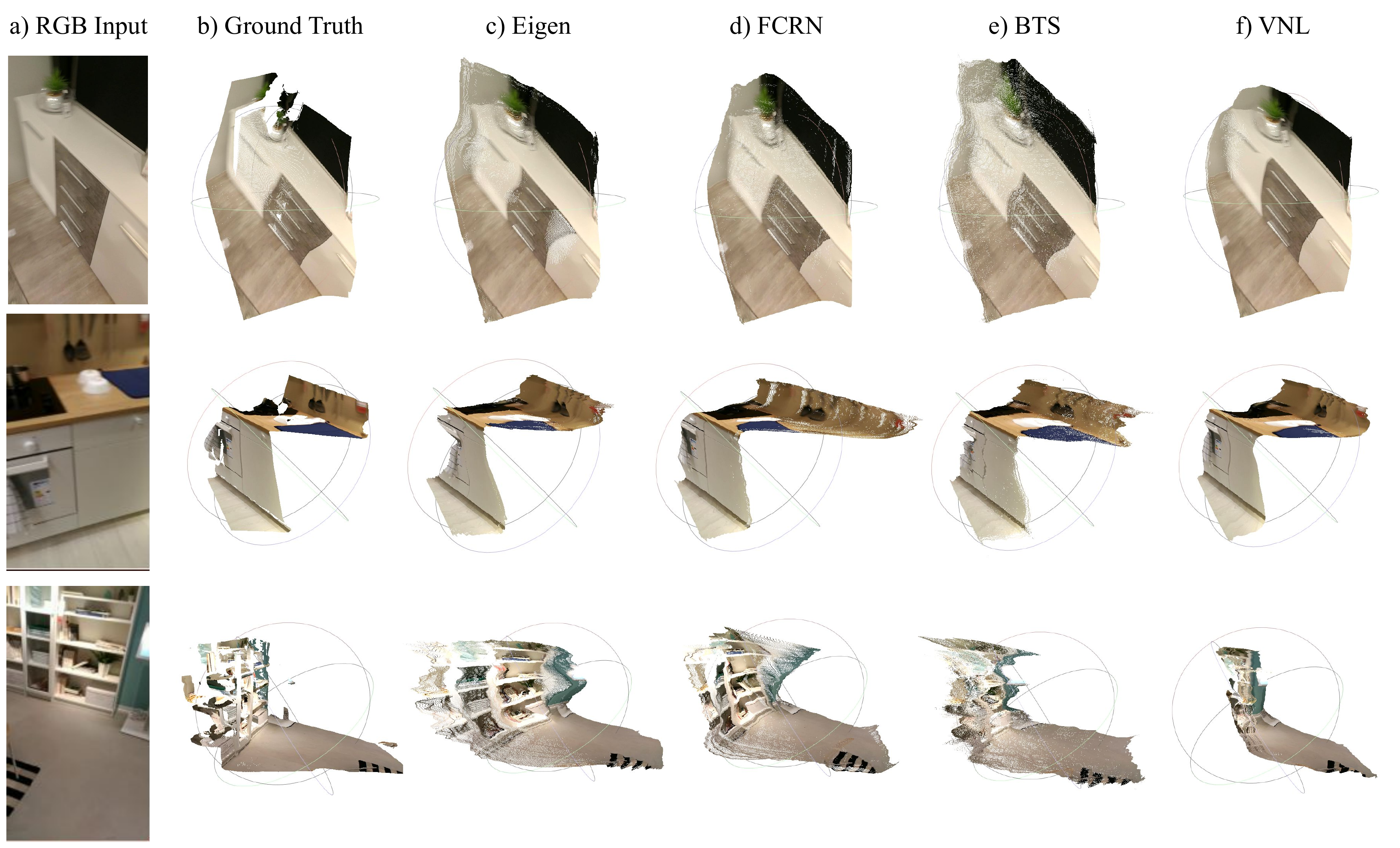}
   \caption{Qualitative depth prediction results on 3 different samples with a) RGB input and corresponding b) ground truth depth, next to the prediction of c) Eigen, d) FCRN, e) BTS and f) VNL.}
\label{fig:qualitative}
\end{figure*}

\subsection{Monocular Depth Prediction Benchmarking}

We provide a benchmark for RIO-D3D through selected MDP methods: Eigen \cite{eigen2014}, FCRN \cite{laina2016}, BTS \cite{lee2019bts} and VNL \cite{yin2019virtualnormal}. When training the models, we follow the original implementations and hyperparameters in terms of batch size, learning rate, input image size and augmentation. 


\noindent\textbf{Evaluation based on 2D metrics} Quantitative results are provided in Tbl. \ref{tab:metrics}.
It can be observed that the standard 2D metrics are saturated and vary with an insignificant difference.
As can be seen in the qualitative results (Fig. \ref{fig:qualitative}) that show point clouds obtained from back-projected predictions, methods cannot predict 3D structures well, and the standard 2D metrics do not capture this. 
The metric scores do not align well with the observed prediction quality. 
Despite the good metric scores, BTS prediction point clouds are full of outliers and fail to capture the 3D object surface. 

This highlights the need for methods and metrics which focus on 3D aspects to better quantify the quality of predictions that aligns with visual observations.

\noindent\textbf{Evaluation based on 3D metrics} Eigen \cite{eigen2014} and FCRN \cite{laina2016} have same absrel scores with $0.21$, however, FCRN is better in terms of 3D metrics which is also evidenced by point cloud visualizations on Fig. \ref{fig:qualitative}. Likewise, a recent method BTS lies within the similar range on 2D metrics with older methods Eigen and FCRN, yet again, as it can be seen on the point clouds, it is structurally superior than others, which is also indicated with the best F-score on the Tbl. \ref{tab:metrics}. When it comes to VNL, it performs worse than the other methods. The standard metrics show only a slight change of $\%16$ ($0.25$ absrel compared to $0.21$ FCRN absrel), whereas 3D metrics reflect the quality better by showing a higher change $\%35$  ($0.76$ CD compared to $0.49$ FCRN CD).

\begin{figure}[t]
    \centering
    \includegraphics[width=\linewidth]{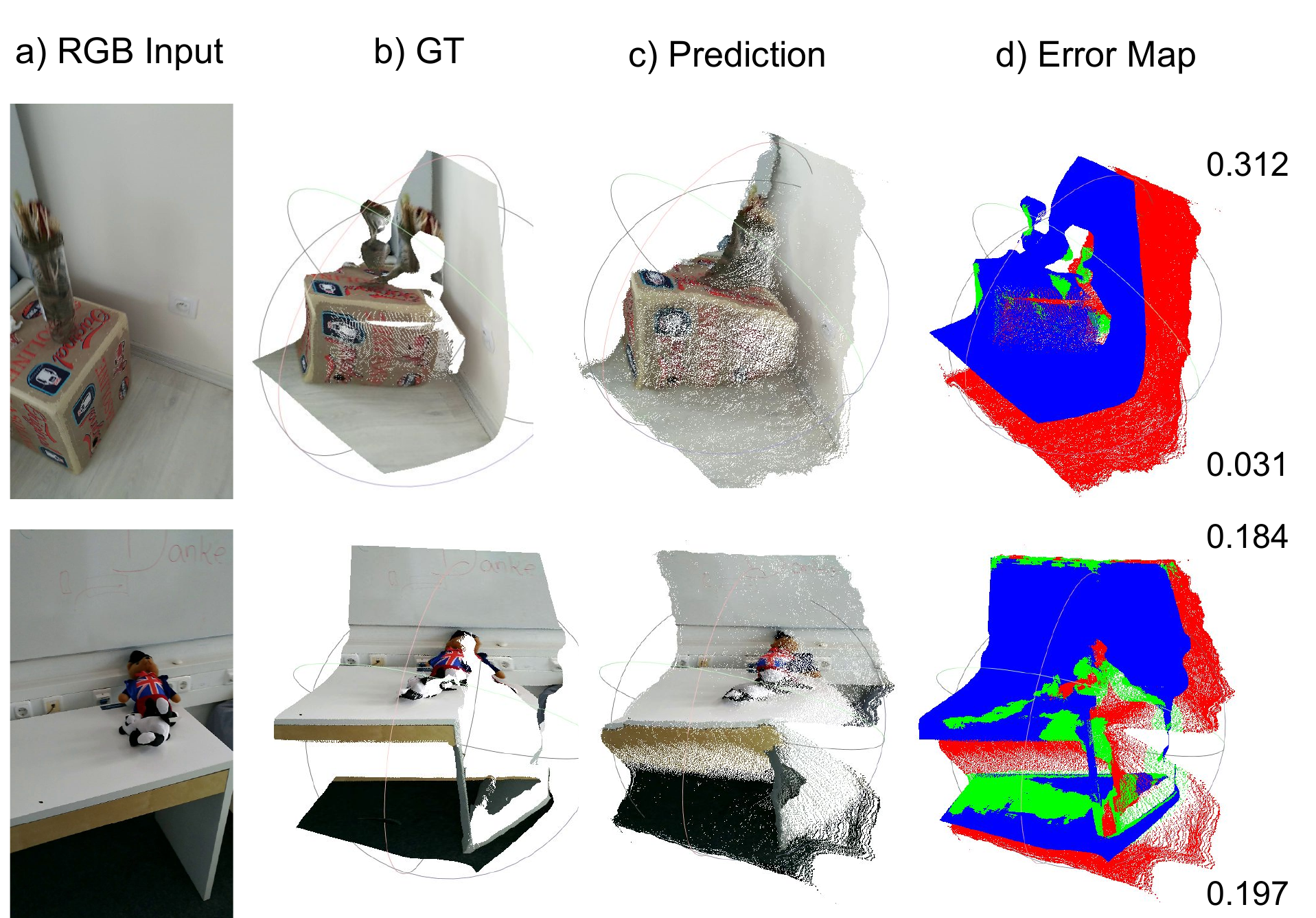}
   \caption{Predictions with relatively good absrel $\downarrow$ (top right) and bad F-score $\uparrow$ (bottom right). Error map (f) colors indicate \color{blue}{$\blacksquare$} \color{black}ground truth \color{red}{$\blacksquare$} \color{black}prediction \color{green}{$\blacksquare$} \color{black}intersection.}
\label{fig:metrics_comparison}
\end{figure}

\begin{figure*}[t!]
    \centering
        \includegraphics[width=0.9\linewidth]{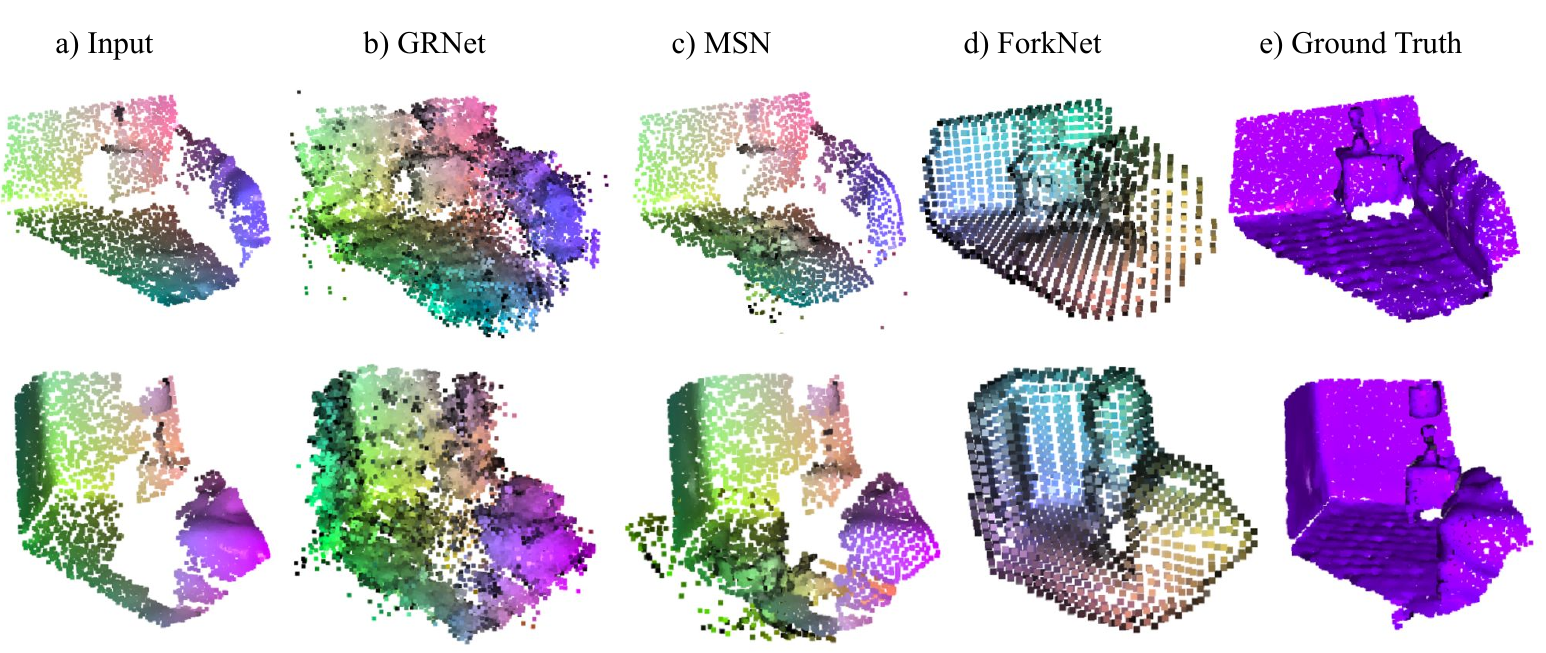}
    \caption{Scene completion results with 2,048 input partial point cloud, GRNet~\cite{grnet_xie}, MSN~\cite{liu2020morphing},  ForkNet~\cite{wang2019forknet}, complete ground truth point cloud.}
    \label{fig:semantic_completion}
\end{figure*}

\noindent\textbf{Comparing 2D metrics to 3D metrics} The current implementations of 3D based metrics follow the trend of standard metrics closely (as in Tbl \ref{tab:metrics}). If one method is the best on standard metrics, it is also the comparably better on 3D metrics. Yet, when looking at the individual samples, there are numerous conflicting examples, \textit{i.e.}, good absrel value but bad F-score or vice versa, as in Figures \ref{fig:teaser}(a,b) and \ref{fig:metrics_comparison}. Note that the first sample in Fig. \ref{fig:metrics_comparison} has an absrel value comparable to the test average but a poor F-score. Similarly, the second sample with an absrel that is on-par with test average, however, has a poor 3D structure, which is also reflected by the F-score. 

The major advantage of the 3D metrics is that they align better with observed qualitative results and quantify the quality of the structures present in the scene more accurately. Even when the intersection between ground truth and prediction is small, absrel values show low variance. The F-score values however quickly get worse when structural overlap is reduced. 2D metrics evaluate averaged per pixel depth error which can be small even if the depth is highly inaccurate. For poor absrel, the 3D metrics are also poor. However, for a small range of absrel values, 3D scores have a larger variety and better reflect the quality of 3D structure. 

\noindent\textbf{Comparison between 3D metrics} CD and EMD are commonly used for evaluating 3D shape reconstruction. Yet, they depend on k-nearest neighbour matching which can lead to unreliable scores due to outliers. Comparably, point cloud based IoU and F-score are more robust, because even when the knn matching is wrong, IoU and F-score do not magnify the effect through falsely identified distances. They show the quality of 3D shapes more clearly, \textit{e.g.}, the interpolation effect on occlusion boundaries and poor planar surfaces. The range of IoU/F-score values depend on the selected threshold, in our experiments set as 0.01 meters, and smaller thresholds ($\leq 10 cm$) are too strict given potential noise levels of depth sensors and the ground truth. IoU is more sensitive than F-score and gives lower values for poor depth maps, though beyond a threshold, results in sudden high spikes. 

\noindent\textbf{Recommended metric} Even though we invite the community to take a critical standpoint, F-score shows to be more advantageous over the others. Calculating the harmonic mean between precision and recall, it measures the percentage of 3D points reconstructed correctly. It is more reliable than CD/EMD/Comp., more robust than IoU, and considers both the structural quality and the amount of correctly retrieved structures. Hence, F-score is the most promising 3D indicator amongst the proposed ones for evaluating MDP. 

\subsection{Scene Completion Benchmarking}


RIO-D3D comes with the availability of full 3D scenes along with the partial depth maps. This can serve as benchmark for the 3D single-view scene completion \cite{yang2018foldingnet, wang_softpool, grnet_xie}, an active research field which recently witnessed a lack of benchmarking tools due to the difficulty of obtaining completion ground truth and the disappearance of one of the main datasets (SUNCG \cite{song2016suncg}). 
Towards this aim, we extend our benchmark to the completion task, where the partial scene input is based on the depth dataset, and the completed scene input is extracted from the full 3D mesh. We provide a comparison among state of the art scene completion methods in Tbl. \ref{tab:scannet}, where a 10-frame qualitative baseline for semantic completion are shown in Fig. \ref{fig:semantic_completion}. VRCNet~\cite{Pan_2021_CVPR} and SoftPoolNet~\cite{wang_softpool} are applied in a cascaded way for 3D completion and semantic segmentation separately. It can be seen that the point-cloud based methods (GRNet~\cite{grnet_xie} and MSN~\cite{liu2020morphing}) produce noisy results as they specifically focus on completion tasks where samples do not vary significantly in both geometries and poses. However, voxel based approaches (ForkNet~\cite{wang2019forknet}) work better both in terms of CD and EMD, and qualitatively. In general, our scenes have a wider variety, more diverse geometry and depths than the existing scene completion datasets.

\begin{table}[t]
\caption{Evaluation of the scene completion with CD ($\times10^3$) and EMD ($\times10^4$) on RIO-D3D.}
\centering
\begin{tabular}{p{3cm}|c|c}
\toprule
\multicolumn{1}{l}{\textbf{Method}}
& CD$\downarrow$ & EMD$\downarrow$ \\
\midrule
FoldingNet~\cite{yang2018foldingnet} & 18.64 & 11.78 \\
PCN~\cite{yuan2018pcn} & 13.18 & 10.63 \\
MSN~\cite{liu2020morphing} & 8.72 & 5.31 \\
SoftPoolNet~\cite{wang_softpool} & 10.07 & 5.76 \\
GRNet~\cite{grnet_xie} & 7.55 & 5.20 \\
VRCNet~\cite{Pan_2021_CVPR} & \textbf{6.24} & \textbf{4.92} \\
\bottomrule
\end{tabular}
\label{tab:scannet}
\vspace{-1.5em}
\end{table}

\section{CONCLUSION}

It is crucial to rethink the MDP task regarding structural accuracy rather than a global pixel-wise 2D error because a depth map, though represented in 2D format, actually consists of 3D information about the environment. As current research fails to address this issue suitably, more appropriate methods, losses, and metrics are imperative. We aim to enable novel ideas to capture scene geometry finer than existing methods. To this end, we point out 3D metrics that directly evaluate the 3D structures of depth maps and provide a suited depth prediction benchmark with complete 3D structures and high-quality depth maps. Our experiments show F-score is the most promising indicator. RIO-D3D and our evaluation codes will be publicly available. Having brought this forth, we hope that future research on MDP will do better than retrieving planes and images.



\bibliography{egbib}

\end{document}